\documentclass{nle}

\usepackage{booktabs} 
\usepackage{url}
\usepackage{amsmath}
\usepackage{mathtools}
\usepackage[T1]{fontenc}
\usepackage{graphicx}
\usepackage[ruled]{algorithm2e} 

\begin{document}
\title[Unsupervised Separation of Native and Loanwords for Malayalam and Telugu]{Unsupervised Separation of Native and Loanwords for Malayalam and Telugu\footnote{Manuscript currently submitted to \emph{Natural Language Engineering} for peer review.}\footnote{This is an extended version of a conference paper (Ref: https://arxiv.org/abs/1803.09641) that has been enriched with substantive new content, with significant extensions on both the method modeling and the experiments.}}

\author[Sridhama Prakhya and Deepak P]{Sridhama Prakhya$^1$ and Deepak P$^2$ \\
             $^1$HHMI Janelia Research Campus, Ashburn VA, USA \\
             $^2$Queen's University Belfast, UK \\
             $^1$sridhama@sridhama.com $^2$deepaksp@acm.org}

\maketitle

\begin{abstract}

Quite often, words from one language are adopted within a different language without translation; these words appear in transliterated form in text written in the latter language. This phenomenon is particularly widespread within Indian languages where many words are loaned from English. In this paper, we address the task of identifying loanwords automatically and in an unsupervised manner, from large datasets of words from agglutinative Dravidian languages. We target two specific languages from the Dravidian family, viz., Malayalam and Telugu. Based on familiarity with the languages, we outline an observation that native words in both these languages tend to be characterized by a much more versatile stem - stem being a shorthand to denote the subword sequence formed by the first few characters of the word - than words that are loaned from other languages. We harness this observation to build an objective function and an iterative optimization formulation to optimize for it, yielding a scoring of each word's nativeness in the process. Through an extensive empirical analysis over real-world datasets from both Malayalam and Telugu, we illustrate the effectiveness of our method in quantifying nativeness effectively over available baselines for the task.


\end{abstract}
\section{Introduction}

Malayalam and Telugu are two widely spoken languages in southern India: Malayalam is an official state language of Kerala, Lakshadweep, and Mahe while Telugu is the official state language of Telangana and Andhra Pradesh. Malayalam is spoken by 37 million native speakers, whereas Telugu has 70 million native speakers\footnote{http://www.vistawide.com/languages/top\_30\_languages.htm}. Both languages are agglutinative and come under the Dravidian language family. Agglutinative languages are characterized by the flexibility they offer to form complex words by chaining simpler morphemes together. The growing web presence of these languages necessitates automatic techniques to process text in them. It is estimated that Indian language internet users will exceed the English user base by 2021\footnote{http://bestmediainfo.com/2018/01/regional-language-users-to-account-for-75-of-total-internet-users-by-2021-times-internet-study/}, underlining the importance of developing effective NLP for Indian languages. A hurdle in exploiting the presence of Malayalam and Telugu text from social media to train models for NLP tasks such as machine translation, named entity recognition and POS tagging, is that of the presence of a large number of loanwords within text from these languages. The loanwords are predominantly from English, and many loanwords, such as {\it police}, {\it train} and {\it taxi} virtually always appear in transliterated form in contemporary Malayalam and Telugu texts. On a manual analysis of a Malayalam news dataset, we found that up to 25\% of the vocabulary were formed by loanwords. While processing mixed language text for tasks such as translation or tagging, automatically identifying loanwords upfront and flagging them would help avoid treating them as separate token (wrt their source language versions) directly leading to enhanced effectiveness of the model under the same learning method. The separation of intrinsic language words from loanwords is especially useful in the realm of cross language information retrieval.

In this paper, we consider the task of separating {\it loanwords} from the {\it native} language words within an unlabeled dataset of words gathered from a document corpus in the language of interest (i.e., either Malayalam or Telugu). We propose an {\it unsupervised} method, {\bf U}nsupervised {\bf N}ativeness {\bf S}coring, that takes in a dictionary of Malayalam or Telugu words, and scores each word in the dictionary based on their {\it nativeness}. UNS uses an optimization framework, which starts with scoring each word based on the versatility of its word stem as observed in the corpus, and refines the scoring iteratively leveraging a generative model built over character n-gram probability distributions. Our empirical analysis illustrates the effectiveness of UNS over existing baseline methods that are suited for the task.

\section{Related Work}\label{sec:related}

Identification of loanwords and loanword sequences, being a critical task for cross-lingual text analysis, has attracted attention since the 1990s. While most methods addressing the problem have used supervised learning, there have been some methods that can work without labeled data. We briefly survey both classes of methods. 


\subsection{ Supervised and `pseudo-supervised' Methods}


An early work\cite{chen1996identification} focuses on a sub-problem, that of supervised identification of proper nouns for Chinese. \cite{jeong1999automatic} consider leveraging decision trees to address the related problem of learning transliteration and back-transliteration rules for English/Korean word pairs. Both these and other methods from the same family rely on and require large amounts of training data. Obtaining such amounts of data has high costs associated with it. To alleviate this, \cite{baker2008statistical} propose a rule-based method to generate large amounts of training data for English-Korean loanword identification. Baker and Brew make use of phonological conversion rules to generate training data. They show that a classifier trained on the generated data performs comparably with one trained on actual examples. Although their method makes use of comparatively less manually-labeled training data, it still relies on rules that specify how words change when borrowed. These are not very much applicable for our context of Dravidian languages where words seldom undergo significant structural changes other than in cases involving usage of external sandhis\footnote{https://en.wikipedia.org/wiki/Sandhi} to join them with adjacent words. 




\subsection{Unsupervised Methods}\label{sec:unsupervised}

A recent work proposes that multi-word phrases in Malayalam text where their component words exhibit strong co-occurrence be categorized as transliterable/loanword phrases \cite{prasad2014technique}. Their intuition stems from observing contiguous words such as {\it test dose} which often occur in transliterated form while occurring together, but get replaced by native words in other contexts. Their method is however {\it unable to identify single  loanwords}, or phrases involving words such as {\it train} and {\it police} whose transliterations are heavily used in the company of native Malayalam words. There hasn't been any work, to our best knowledge, on automatically identifying loanwords in Telugu text. However, a recent linguistic study on characteristics of loanwords in Telugu newspapers~\cite{uppuletiloan} is indicative of the abundance and variety of loanwords in Telugu. 


There has been some work that relax supervised data requirements for the task within the context of languages of non-Indic origin. \cite{goldberg2008identification} present a loosely-supervised approach that sources native words from 100-year-old Hebrew texts. The assumption is that there would be fewer foreign words in these older texts. Indian languages, particularly regional south indian languages, are yet to see large-scale digitization efforts of old text for such temporal assumptions to be leveraged in nativeness scoring. \cite{koo2015unsupervised} presents unsupervised loanword identification in Korean where they construct a binary character-based n-gram classifier that is trained on a corpus. Koo makes use of native and foreign seed words that are determined using document statistics. Words with higher corpus frequency are part of the native seed. This is based on the assumption that native words occur more frequently than foreign words in a corpus. The foreign seed consists of words that have apparent vowel insertion. According to Koo, in Korean---as well as phonotactically similar languages---words neither begin nor end with consonant clusters. Therefore, foreign words usually have vowels arbitrarily inserted to break the consonant clusters. Contrary to the phonotactics of Korean, words in Malayalam and Telugu can begin and end with consonant clusters. Koo's method is therefore inapplicable to the languages in our focus.

%
%




\subsection{Positioning the Nativeness Scoring Task} 

Nativeness scoring of words may be seen as a vocabulary stratification step (upon usage of thresholds) for usage by downstream applications. A multi-lingual text mining application that uses Malayalam/Telugu text in the company of English text would benefit by transliterating non-native Malayalam/Telugu words to English, so the loanword token and its transliteration is treated as the same token. For machine translation, loanwords may be channeled to specialized translation methods (e.g.,~\cite{DBLP:conf/acl/TsvetkovD15}) or for manual screening and translation. 

\section{Problem Definition}\label{sec:probdef}

We now define the problem more formally. Consider $n$ distinct words obtained from Malayalam/Telugu text, $\mathcal{W} = \{ \ldots, w, \ldots \}$. It may be noted that $\mathcal{W}$ should either contain all Malayalam words, or all Telugu words (not a mixture of some Telugu and some Malayalam words). This may be obvious for readers familiar with the fact that the two languages use different scripts leading to non-overlapping vocabularies; so, mixing them within a dataset doesn't make much intuitive sense. Our task is to devise a technique that can use $\mathcal{W}$ to arrive at a {\it nativeness} score for each word, $w$, within it, as $w_n \in [0,1]$. 

\begin{equation}
\{ \ldots, w, \ldots \} \xrightarrow[\text{Nativeness Scoring}]{\text{Unsupervised}} \{ \ldots, w_n \ldots \}
\end{equation}

We would like $w_n$ to be an accurate quantification of native-ness of word $w$. Since $w_n \in [0,1]$, $(1-w_n)$ may be treated analogously as a quantification of loanword-ness of $w$. For example, when words in $\mathcal{W}$ are ordered in the decreasing order of $w_n$ scores, we expect to get the native words at the beginning of the ordering and vice versa. We do not presume availability of any data other than $\mathcal{W}$; this makes our method applicable across scenarios where corpus statistics are unavailable due to privacy or other reasons. 

\subsection{Evaluation}\label{sec:evaluationmeasures}

Given that it is easier for humans to crisply classify each word as either a native word or a loanword in lieu of attaching a score to each word, the nativeness scoring (as generated by a scoring method such as ours) often needs to be evaluated against a crisp nativeness assessment, i.e., a scoring with scores in $\{0, 1\}$. Such evaluation involving the comparison of a scoring with a crisp labelling appears in other contexts such as the task of record linkage scoring~\cite{DBLP:conf/pakdd/JurekP18}; within record linkage scenarios, however, the evaluation is further confounded due to the high imbalance between the cardinalities of the two classes in question. \cite{DBLP:conf/pakdd/JurekP18} uses the an aggregate of the rankings of the minority class in an ordering of the objects according to the scores, in order to evaluate the effectiveness of the (record linkage) scoring task. We use a similar framework, but use precision instead of average ranking since the imbalance of sizes between the native and loanword vocabulary is not too extreme in Indian language settings. Consider the ordering of words in the labeled set in the decreasing (or more precisely, non-increasing) order of {\it nativeness} scores (each method produces an ordering for the dataset). We use two sets of metrics for evaluation: 

\begin{itemize}
\setlength\itemsep{0in}
\item {\it Precision at the ends of the ordering:} {\bf Top-k precision} denotes the fraction of {\it native} words within the $k$ words at the {\it top} of the ordering; analogously, {\bf Bottom-k precision} is the fraction of {\it loanwords} among the {\it bottom k}. Since a good scoring would likely put native words at the top of the ordering and the loanwords at the bottom, a good scoring method would intuitively score high on both these metrics. We call the average of the top-k and bottom-k precision for a given k, as {\bf Avg-k precision}. These measures, evaluated at varying values of $k$, indicate the quality of the nativeness scoring at either ends. 
\item {\it Clustering Quality:} Consider the cardinalities of the native and loanword sets from the labeled set as being $N$ and $L$ respectively. We now take the top-N words and bottom-L words from the ordering generated by each method, and compare against the respective labeled sets as in the case of standard clustering quality evaluation\footnote{\url{https://nlp.stanford.edu/IR-book/html/htmledition/evaluation-of-clustering-1.html}}. Since the cardinalities of the generated native (loanword) cluster and the native (loanword) labeled set is both $N$ ($L$), the {\bf Recall} of the cluster is identical to its {\bf Purity/Precision}, and thus, the {\bf F-measure} too; we simply call it {\bf Clustering Quality}. A cardinality-weighted average of the clustering quality across the native and loanword clusters yields a single value for the clustering quality across the dataset. It may be noted that, as expected, we are not making the labeled dataset available to the method generating the ordering, instead merely using it's cardinalities for evaluation purposes. 
\end{itemize}


%


\section{UNS: Unsupervised Nativeness Scoring}\label{sec:method}

We now introduce our method, {\bf U}nsupervised {\bf N}ativeness {\bf S}coring. We use probability distributions over character n-grams to separately model loanwords and native words, and develop an optimization framework that alternatively refines the character n-gram distributions and nativeness scoring within each iteration. UNS involves an initialization that induces a {\it coarse} separation between native and loanwords followed by iterative refinement. The initialization is critical in optimization methods that are vulnerable to local optima; the native word distribution needs to be initialized to  {\it roughly} prefer native words over loanwords. This will enable further iterations to exploit the initial preference direction to further refine the model to {\it attract} the native words more strongly and weaken any initial preference to loanwords. The vice versa holds for the models that stand for loanwords. We will first outline the initialization step followed by the description of the iterative framework and the overall approach. UNS, it may be noted, is designed keeping Malayalam and Telugu in mind; thus, UNS can be applied on an input dictionary $\mathcal{W}$ comprising either Malayalam {\bf or} Telugu words, but not a mix of words from Malayalam and Telugu. 


%

\subsection{Diversity-based Initialization}\label{sec:divinit}


Our initialization is inspired by an observation of the versatility of word stems. We define a word stem as the sub-word formed by the first few pseudo-syllables of a (Malayalam or Telugu) word; a pseudo-syllable is one which contains a consonant along with one or more modifiers that appear with it. Consider a word stem {\it |pu|ra|}\footnote{{\scriptsize We will represent Malayalam/Telugu words in transliterated form for reading by those who might not be able to read Malayalam/Telugu. A pipe would separate Malayalam/Telugu pseudo-syllable; a pseudo-syllable is one which contains a consonant along with one or more modifiers that appear with it, modifiers being primarily those listed in https://www.win.tue.nl/~aeb/natlang/malayalam/malayalam-alphabet.html. for example |pu| corresponds to a character /pa/ along with a modifier /u/. }}, a stem commonly leading to native Malayalam words; its suffixes (i.e., subwords that could immediately follow them to form a full word) are observed to start with a variety of characters such as {\it |ttha|} (e.g., {\it |pu|ra|ttha|kki|}), {\it |me|} (e.g., {\it |pu|ra|me|}), {\it |mbo|} (e.g., {\it |pu|ra|mbo|kku|}) and {\it |ppa|} (e.g., {\it |pu|ra|ppa|du|}). On the other hand, stems that mostly lead to loanwords often do not exhibit so much of diversity. For example, {\it |re|so|} is followed only by {\it |rt|} (i.e., {\it resort} being a commonly used loanword from English) and {\it |po|li|} is usually only followed by {\it |s|} (i.e., {\it police}). Some stems such as {\it |o|ppa|} lead to transliterations of two English-origin loanwords such as {\it opener} and {\it operation}. To sum up, our observation, upon which we model the initialization part of UNS, is that the {\it variety of suffixes is generally correlated with native-ness (i.e., propensity to lead to a native word) of word stems}. This is intuitive since loanword stems, being of non-native origin, would be expected to provide limited versatility to being modified by sandhis or derivational/inflectional suffixes as compared to native ones. 


For simplicity, we use the first two pseudo-syllables (characters grouped with their modifiers) of each word as the word stem; we will evaluate the robustness of UNS to varying stem lengths in our empirical evaluation, while consistently using the stem length of two pseudo-syllables in our description. We start by associating each distinct word stem in $\mathcal{W}$ with the number of unique third pseudo-syllables that follow it (among words in $\mathcal{W}$); in our examples, {\it |pu|ra|} and {\it |o|pa|} would be associated with $4$ and $2$ respectively. We initialize the {\it nativeness} weights as proportional to the diversity of 3$^{rd}$ pseudo-syllables beyond the stem:

\begin{equation}\label{eq:init}
w_{n_{0}} = min\bigg\{ 0.99, \frac{|u3(w_{stem},\mathcal{W})|}{\tau}\bigg\}
\end{equation}

\noindent where $u3(w_{stem},\mathcal{W})$ denotes the set of third pseudo-syllables that follow the stem of word $w$ among words in $\mathcal{W}$. We flatten off $w_{n_0}$ scores beyond a diversity of $\tau$ (note that a diversity of $\tau$ or higher will lead to the second term in the expression above becoming $1.0$ or higher, kicking in the min function to choose $0.99$ for $w_{n_0}$) as shown in the above equation. By disallowing $w_{n_0}$ to assume the maximum possible value of $1.0$, we will allow even words formed of highly versatile stems to contribute, albeit very slightly, to building loanword pseudo-syllable n-gram models (details of which are in the next section); this limits over-reliance on the versatility initialization heuristic. We set $\tau = 10$ based on our observation from the dataset that most word stems having more than $10$ distinct pseudo-syllables were seen to be native. As in the case of word stem length, we study UNS trends across varying $\tau$ in our empirical analysis. 


\subsection{UNS Iterative Optimizations}

Having arrived at an initialization of nativeness scores $\{ \ldots, w_{n_0}, \ldots \}$, UNS refines the scores iteratively in order to arrive at an accurate quantification of nativeness. We use characters in the below narrative to refer to pseudo-syllables (the former being more familiar terminology); pseudo-syllables may either be single characters, or characters grouped with their modifiers. The UNS structure of iterative refinement makes use of two models, which we first introduce:

\begin{itemize}
\item {\bf Native and Loanword Character n-gram Distributions:} Probability distributions over character n-grams form the main tool used within UNS to refine the word nativeness scores. UNS uses separate probability distributions over character n-grams to model native and loanwords. While the size of the n-grams (i.e., whether $n=1,2,3$ or $4$) over which the probability distributions are built is a system-level parameter, we will use $n=1$ for simplicity in our description of UNS. We denote the native and loanword character probability distributions as $\mathcal{N}$ and $\mathcal{L}$ respectively, with $\mathcal{N}(c)$ and $\mathcal{L}(c)$ denoting the weight associated with the character (1-gram) $c$ according to the respective distributions. 
\item {\bf Highly Diverse Words:} UNS works by refining the initialization through iterations in order to arrive at an accurate nativeness quantification for words in $\mathcal{W}$. Over the course of iterations, the n-gram distribution-based model could drag the $w_n$ scores afar from their initialized values. UNS uses a mechanism in order to ensure that there is an inertia to such movements in the case of words with highly versatile stems. This brings us to the second model in UNS, a subset of words from $\mathcal{W}$ with highly versatile stems (beyond a threshold $\rho$), which we will denote as $\mathcal{W}_D$:
\[ \mathcal{W}_D = \{ w | w \in \mathcal{W} \wedge u3(w_{stem}) > \rho \} \]
Thus, all words formed using stems that are versatile enough to to be followed by more than $\rho$ different characters in $\mathcal{W}$ would form part of $\mathcal{W}_D$. 
\end{itemize}

These models are leveraged in an iterative optimization framework. Towards deriving the optimization steps, we first outline two objective functions in the next section. 

\subsubsection{Maximizing and Minimizing Objective Functions}

Consider an estimate of the nativeness scores $\{ \ldots, w_n, \ldots \}$ for all words in $\mathcal{W}$, and a state of the character unigram models $\mathcal{N}$ and $\mathcal{L}$. Consider a particular word $w \in \mathcal{W}$; if it has a high nativeness (non-nativeness) score, we would reasonably expect $w$ to be formed by characters that score highly within the $\mathcal{N}$ ($\mathcal{L}$) character probability distribution. This can be folded into an intuitive objective function that would be maximized under a good estimate of word scores and models:

\begin{equation}\label{eq:maximizing1}
\prod_{w \in \mathcal{W}} \prod_{c \in w} \bigg( w_n^2 \times \mathcal{N}(c) + (1-w_n)^2 \times \mathcal{L}(c) \bigg)
\end{equation}

This measures the aggregate supports for words in $\mathcal{W}$, the support for each word measured as an interpolated support from across the distributions $\mathcal{N}$ and $\mathcal{L}$ with weighting factors being squares of the nativeness scores (i.e., $w_n$s) and loanword-ness scores (i.e., $(1-w_n)$s) respectively. In a way, the objective function can be regarded as the likelihood of words being generated by a generative process where words are formed by sampling characters from $\mathcal{N}$ and $\mathcal{L}$ in rates directly and inversely related to the word nativeness score $w_n$ respectively. Similar mixing models have been used earlier in emotion lexicon learning~\cite{bandhakavi2014generating} and solution post discovery~\cite{deepak2014unsupervised}. The squares of the nativeness/loanwordness scores are used in our model (instead of the raw scores) for optimization convenience; it may be noted that the usage of squares has a side-effect of the optimizing pushing the nativeness scores towards either ends of the $[0,1]$ spectrum. A highly native word should intuively have a high $w_n$ (nativeness) and a high support from $\mathcal{N}$ and correspondingly low loanword-ness (i.e., $(1-w_n)$) and support from $\mathcal{L}$; a highly non-native word would be expected to have exactly the opposite. Due to the design of Eq.~\ref{eq:maximizing1} in having the higher terms multiplied with each other (and so for the lower terms), this function would be maximized for a desirable estimate of the variables $\theta = \{ \mathcal{N}, \mathcal{L}, \{ \ldots, w_n, \ldots \} \}$. 

As indicated earlier, in addition to measuring and optimizing for conformance of $\mathcal{N}$ and $\mathcal{L}$ models with nativeness scores, we would like to penalize successive iterations from dragging words in $\mathcal{W}_D$, those being words having highly versatile stems, into a low $w_n$ territory. A simple objective to maximize $w_n$s of words in $\mathcal{W}_D$ would be as follows:

\begin{equation}\label{eq:maximizing2}
\prod_{w \in \mathcal{W}_D} w_n^2
\end{equation}

We put the expressions from Eq.~\ref{eq:maximizing1} and Eq.~\ref{eq:maximizing2} together to form a composite objective function as follows:

\begin{equation}\label{eq:maximizing}
\mathcal{O}_{max} = \bigg[ \prod_{w \in \mathcal{W}} \bigg( \prod_{c \in w} \big( w_n^2 \times \mathcal{N}(c) +  (1-w_n)^2 \times \mathcal{L}(c) \big) \bigg] \bigg[ \prod_{w \in \mathcal{W}_D} w_n^2 \bigg]^{\alpha}
\end{equation}

The parameter $\alpha$ enables controlling the relative weighting for the model conformance and diverse words' inertia terms respectively. At $\alpha = 0$, the function simply becomes the model conformance term, whereas very high values of $\alpha$ would lead to $\mathcal{O}_{max}$ being largely influenced by the second term. The suffix in $\mathcal{O}_{max}$ indicates that the objective function is one that needs to be maximized. 

\noindent {\bf Minimizing Objective:} We now define an analogous construction of an objective function, whose minimization would lead to improving model conformance (with current estimates of $w_n$s) and diverse words' inertia; this is in contrast with $\mathcal{O}_{max}$ for whom higher values indicate better model conformance. The minimizing objective is as follows:

\begin{equation}\label{eq:minimizing}
\mathcal{O}_{min} = \bigg[ \prod_{w \in \mathcal{W}} \bigg( \prod_{c \in w} \big( (1-w_n)^2 \times \mathcal{N}(c) +  w_n^2 \times \mathcal{L}(c) \big) \bigg] \bigg[ \prod_{w \in \mathcal{W}_D} (1-w_n)^2 \bigg]^{\alpha}
\end{equation}

Let us first consider the model conformance term; in this form, given a good estimate of the models, the highly native (non-native) words have their nativeness (loanword-ness) weights multiplied with the support from the loanword (native) character n-gram probability distribution. In other words, maximizing the model conformance term in Eq.~\ref{eq:maximizing1} is semantically equivalent to minimizing the first term in Eq.~\ref{eq:minimizing} above. Similar is the case with the diverse words' inertia term; minimizing the product of $(1-w_n)^2$s is semantically equivalent to maximizing the product of $w_n^2$s (recollecting that $w_n \in [0,1]$). The $\alpha$ parameter, as before, allows to tune the relative weighting between the two terms in the composite objective. 

\noindent {\bf Role of two Objectives:} We have outlined two objective functions above to measure the goodness of the estimates of $w_n$s, $\mathcal{N}$ and $\mathcal{L}$. In UNS, we will optimize for the estimates (i.e., $w_n$s) and models (i.e., $\mathcal{N}$ and $\mathcal{L}$) in alternative steps. The construction of the formulation, given the interpolation of supports from the models, is such that it is analytically difficult to arrive at a formulation to use the same objective function (either $\mathcal{O}_{max}$ or $\mathcal{O}_{min}$) for both optimization steps. Accordingly, we will outline a optimization method that uses the maximizing objective, $\mathcal{O}_{max}$ to optimize for the estimates of the models (i.e., $\mathcal{N}$ and $\mathcal{L}$), while the minimizing objective, $\mathcal{O}_{min}$, is used to optimize for the word nativeness scores, i.e., the $w_n$s. 

\subsubsection{Estimating Word Nativeness Scores}

Our task, in this phase, is to use the current estimates of $\mathcal{N}$ and $\mathcal{L}$ in order to identify a set of nativeness scores that best conform to the models and the inertia term. As indicated earlier, we will use the minimizing objective, $\mathcal{O}_{min}$ in order to estimate the $w_n$s. First, writing out $\mathcal{O}_{min}$ in log form gives the following:

\begin{equation}
\mathcal{O}'_{min} = \bigg[ \sum_{w \in \mathcal{W}} \bigg( \sum_{c \in w} ln\big((1-w_n)^2 \times \mathcal{N}(c) +  w_n^2 \times \mathcal{L}(c) \big) \bigg) \bigg] + 2 \alpha \bigg[ \sum_{w \in \mathcal{W}_D} ln(1-w_n) \bigg]
\end{equation}

Noting that the $w_n$s of each word is nicely segregated into a separate term within the summation, the slope of this objective with respect to the nativeness score of a particualr word $w'$ is as follows:

\begin{equation}\label{eq:slopemin}
\frac{\partial \mathcal{O}'_{min}}{\partial w'_n} = \sum_{c \in w'} \frac{2 w'_n \times (\mathcal{N}(c) + \mathcal{L}(c)) - 2 \mathcal{N}(c)}{(1-w'_n)^2 \times \mathcal{N}(c) +  (w'_n)^2 \times \mathcal{L}(c)} - \frac{2 \alpha \times \mathcal{I}(w' \in \mathcal{W}_D)}{1-w'_n}
\end{equation}

where $\mathcal{I}(.)$ is an indicator function that returns $1.0$ if the internal expression evaluates to true, and $0.0$ otherwise. An optimal value of $w'_n$ may be achieved by identifying a value of $w'_n$ that brings the slope above to $0.0$. While we omit details here, the second derivative of $\mathcal{O}'_{min}$, when worked out, leads to a positive value, indicating that the equating the slope to $0.0$ would lead to a minima (as against a maxima). Equating Eq.~\ref{eq:slopemin} to $0.0$ and solving for $w'_n$ gives:

\begin{equation}\label{eq:updatewn}
w'_n = \frac{\frac{\alpha \times \mathcal{I}(w' \in \mathcal{W}_D)}{1-w'_n} + \sum_{c \in w'} \frac{\mathcal{N}(c)}{(1-w_n)^2 \mathcal{N}(c) + (w'_n)^2 \mathcal{L}(c)}}{\sum_{c \in w'}\frac{\mathcal{N}(c) + \mathcal{L}(c)}{(1-w'_n)^2 \times \mathcal{N}(c) +  (w'_n)^2 \times \mathcal{L}(c)}}
\end{equation}

It may be seen that Eq.~\ref{eq:updatewn} above is {\it not in a closed form}, since the estimate of $w'_n$ depends on itself, given that it appears in the right-hand-side of the Eq.~\ref{eq:updatewn}. Nevertheless, it offers a constructive way of estimating new values of $w'_n$s by using previous estimates in the right-hand-side of the equation. The numerator of Eq.~\ref{eq:updatewn} comprises two terms; ignoring the first term, it is easy to observe that words comprising characters that score highly in $\mathcal{N}$ (notice that $\mathcal{N}(c)$ appears in the numerator whereas the analogous term in the denominator is $\mathcal{N}(c) + \mathcal{L}(c)$) would achieve high $w'_n$ scores. This formulation, leading to estimating $w'_n$ as being roughly proportional to the support from $\mathcal{N}$ is intuitively desirable. Now, coming to the first term in the numerator, it may be observed that it evaluates to $0.0$ for words not belonging to $\mathcal{W}_D$; for those words in the highly diverse list, it translates into a slight `nudge', one that would push the score slightly upward, once again a desirable effect given that we want to retain a high nativeness score for words in $\mathcal{W}_D$. 

\subsubsection{Learning $\mathcal{N}$ and $\mathcal{L}$ Distributions}

As outlined earlier, we use separate character n-gram probability distributions to model native and loanwords. We would like these probability distributions to support the latest estimates of nativeness scoring and loanwordness scoring respectively. While refining $\mathcal{N}$ and $\mathcal{L}$, we would like to ensure that they remain true probability distributions that sum to unity. This brings in the following constraints:

\begin{equation}
\sum_{c} \mathcal{N}(c) = \sum_{c} \mathcal{L}(c) = 1.0
\end{equation}

We will use the maximizing objective in order to optimize for the $\mathcal{N}$ and $\mathcal{L}$ models. As earlier, taking the log form of the objective and adding Lagrangian terms from the constraints yields the following objective to maximize:

\begin{multline}\label{eq:loglagrangian}
\mathcal{O}'_{max} = \sum_{w \in \mathcal{W}} \bigg( \sum_{c \in w} ln\big(w_n^2 \times \mathcal{N}(c) + (1-w_n)^2 \times \mathcal{L}(c) \big) \bigg) + 2 \alpha \sum_{w \in \mathcal{W}_D} ln(w_n) + \\ \lambda_{\mathcal{N}} \sum_{c} \mathcal{N}(c) + \lambda_{\mathcal{L}} \sum_{c} \mathcal{L}(c)
\end{multline}

The last two terms come from the constraints and are associated with their own Lagrangian multipliers, $\lambda_{\mathcal{N}}$ and $\lambda_{\mathcal{L}}$. 

\noindent {\bf Learning $\mathcal{N}$:} Fixing the values of $w_n$s and $\mathcal{L}$, let us now consider learning a new estimate of $\mathcal{N}$. The slope of Eq.~\ref{eq:loglagrangian} with respect to $\mathcal{N}(c')$, i.e., the weight associated with a particular character $c'$, would be the following:

\begin{equation}
\frac{\partial \mathcal{O}'_{max}}{\partial \mathcal{N}(c')} = \bigg( \sum_{w \in \mathcal{W}} \frac{freq(c',w) \times w_n^2}{w_n^2 \times \mathcal{N}(c') + (1-w_n)^2 \times \mathcal{L}(c')} \bigg) - \lambda_{\mathcal{N}} 
\end{equation}

where $freq(c',w)$ is the frequency of character $c'$ in the word $w$ and $\lambda_{\mathcal{N}}$ denotes the Lagrangian multiplier corresponding to the sum-to-unity constraints for $\mathcal{N}$. Equating this to zero, as earlier, {\it does not yield a closed form solution} for $\mathcal{N}(c')$, but a simple re-arrangement yields an iterative update formula:

\begin{equation}\label{eq:updaten}
\mathcal{N}(c') \propto \sum_{w \in \mathcal{W}} \frac{freq(c',w) \times w_n^2}{w_n^2 + (1-w_n)^2 \times \frac{\mathcal{L}(c')}{\mathcal{N}(c')}}
\end{equation}

As in Eq.~\ref{eq:updatewn}, the previous estimate of $\mathcal{N}(c')$ would need to be used in the right-hand-side of the update equation. The second derivative of $\mathcal{O}'_{max}$ yields a negative value in the general case, this affirming that equating slope to $0.0$ yields a maxima (as against a minima); we omit details here. It is this contrasting behavior of the second derivatives of $\mathcal{O}'_{min}$ and $\mathcal{O}'_{max}$ that requires us to use these two separate objectives for estimating the nativeness scores and probability distributions respectively. Eq.~\ref{eq:updaten} may be seen to be intuitively reasonable, with it establishing a somewhat direct relationship between $\mathcal{N}$ and $w_n$, allowing words with high nativeness to influence $\mathcal{N}$ more. The sum-to-unity constraint can be factored in by simply using the above relation as an update equation, followed by updating the revised estimate using a normalization as follows (this is the same process as in estimating simple maximum likelihood language models):

\begin{equation}\label{eq:normalizen}
\mathcal{N}(c') = \frac{\mathcal{N}(c')}{\sum_c \mathcal{N}(c)}
\end{equation}

\noindent {\bf Learning $\mathcal{L}$:} In a sequence of steps analogous to that as for $\mathcal{N}$ above, we arrive at the following update equation:

\begin{equation}\label{eq:updatel}
\mathcal{L}(c') \propto \sum_{w \in \mathcal{W}} \frac{freq(c',w) \times (1-w_n)^2}{(1-w_n)^2 + w_n^2 \times \frac{\mathcal{N}(c')}{\mathcal{L}(c')}}
\end{equation}

Treating the above proportionality as an equation helps arriving at an update formula, which would then be followed by a normalization on the lines of Eq.~\ref{eq:normalizen}. Analogous to the case of Eq.~\ref{eq:updaten}, Eq.~\ref{eq:updatel} establishes a direct relationship between $\mathcal{L}(c')$ and the loanwordness (i.e., $(1-w_n)$ scores) of words within which $c'$ occurs with high frequency. 

\begin{algorithm}
\DontPrintSemicolon 
\KwIn{A set of Malayalam words or Telugu words, $\mathcal{W}$}
\KwOut{A nativeness scoring $w_n \in [0,1]$ for every word $w$ in $\mathcal{W}$}
{\bf Parameters/Hyper-parameters:} word stem length, initialization diversity threshold $\tau$, size of character n-grams $n$, the threshold to determine highly diverse words $\rho$, the relative weighting in optimization $\alpha$ \;
Initialize the $w_n$ scores for each word using the diversity metric in Eq.~\ref{eq:init} using the chosen stem length and $\tau$\;
\While{\text{not converged and number of iterations} \text{not reached}}{
  \text{Estimate n-gram distributions} $\mathcal{N}$ and $\mathcal{L}$ using Eq.~\ref{eq:updaten} and Eq.~\ref{eq:updatel} respectively\;
  \text{Normalize} the $\mathcal{N}$ and $\mathcal{L}$ distributions so they sum to unity \;
  \text{Learn nativeness weights for each word} \text{using Eq.~\ref{eq:updatewn}} \;
}
\Return{latest estimates of nativeness weights}\;
\caption{{\sc UNS}}
\label{algo:uns}
\end{algorithm}

\subsection{The UNS Iterative Refinement Algorithm}

Having described the initiatalization and the details of the iterative refinement, we now outline the overall UNS algorithm as Algorithm~\ref{algo:uns}. The iterative method starts with the diversity-based initialization, and followed on by a number of iterations, each iteration involving estimation of $\mathcal{N}/\mathcal{L}$ followed by the $w_n$s. Since we do not have closed form solutions for these updates, we use the iterative update steps as outlined earlier. The iterations are stopped when the nativeness weights do not change significantly (as estimated by a threshold) or when a reasonable number of iterations have been completed (we use $100$). It may be noted that the character n-gram distributions within $\mathcal{N}/\mathcal{L}$ may not necessarily be unigrams since $n$ is a parameter to the method; unigrams corresponds to a choice of $n=1$. For $n=2$, the update steps would need to have the inner summations iterating over 2-length character sequences instead of characters; this involves replacing $c \in w$ with $[c_i,c_j] \in w$ in each of the update equations where $[c_i,c_j]$ denotes a contiguous sequence of two characters. Each of the update steps in UNS are linear in the size of $\mathcal{W}$, making UNS a fast technique for even large dictionaries. 

\section{Experiments}\label{sec:expts}
We now describe our empirical study of UNS, starting with the dataset and experimental setup leading on to the results and analyses.

\subsection{Datasets}\label{sec:datasets}
%

\begin{table*}
\centering
 \caption{Frequency of native and loanwords in datasets}
 \label{tab:word_freq}
\begin{tabular}{l|cc}
\toprule
Language & Native & Non-Native \\
\midrule
Malayalam & 777 & 258 \\
Telugu & 778 & 257 \\
\bottomrule
\end{tabular}
\end{table*}

Given the design of our task, we create two separate datasets for our target languages viz. Malyalam and Telugu. Both datasets were created in a similar fashion by sourcing words from articles found in popular newspapers in the respective languages: 65068 unique Malayalam words were obtained from {\it Mathrubhumi}\footnote{{\scriptsize \url{https://www.mathrubhumi.com/}}}, while 43150 distinct Telugu words were sourced from {\it Andhrabhoomi}\footnote{{\scriptsize \url{http://www.andhrabhoomi.net/}}}. For each language, we chose a subset of 1035 random words to be manually labelled as either {\it native}, {\it loanword} or {\it unknown}; this forms our evaluation set. The complete word lists along with the annotated subsets have been made publicly available\footnote{{\scriptsize Malayalam dataset: \url{https://goo.gl/DOsFES}}} \footnote{{\scriptsize Telugu dataset: \url{https://goo.gl/xsvakx}}} \footnote{{\scriptsize Malayalam labeled subset: \url{https://goo.gl/XEVLWv}}} \footnote{{\scriptsize Telugu labeled subset: \url{https://goo.gl/S2eoB2}}}. For evaluation purposes, we merged the set of {\it unknown} labelled words with {\it loanwords}; this seemed appropriate since most {\it unknown} labellings were seen to correlate with {\it non-native} words, whose source language wasn't as obvious as others. The frequencies of native and loanwords in our datasets are shown in Table~\ref{tab:word_freq}. In general, our datasets contain approximately 3 times as many native words as loanwords. This is in tandem with the contemporary distribution of words in the target languages within the news domain, as observed from other sources as well.

\subsection{Baselines}

As outlined in Section~\ref{sec:related}, the unsupervised version of the problem of telling apart native and loanwords for Malayalam and/or similar languages has not been addressed in literature, to the best of our knowledge. The unsupervised Malayalam-focused method\cite{prasad2014technique} (Ref: Sec~\ref{sec:unsupervised}) is able to identify only contiguous sequences of two or more loanwords, making it inapplicable for general contexts where individual english words are often transliterated for want of a suitable malayalam alternative. As an example,~\cite{prasad2014technique} would be able to identify {\it police} as a loanword only in cases where it appears together with other words; while such scenarios, such as {\it police station} and {\it traffic police} do appear many a time, {\it police} abundantly appears in itself. The Korean method\cite{koo2015unsupervised} is too specific to Korean language and cannot be used for other languages due to the absence of a generic high-precision rule to identify a seed set of loanwords. With both the unsupervised state-of-the-art approaches being inapplicable for our task, we compare against an intuitive generalization-based baseline, called {\bf GEN}, that orders words based on their support from the combination of a unigram and bi-gram character language model learnt over $\mathcal{W}$; this leads to a scoring as follows:

\begin{equation*}
w_n^{GEN} = \hspace{3in}
\end{equation*}


\begin{equation}\label{eq:gen}
\prod_{[c_i, c_{i+1}] \in w} \lambda \times B_{\mathcal{W}}(c_{i+1}|c_i) + (1-\lambda) \times U_{\mathcal{W}}(c_{i+1})
\end{equation}

where $B_{\mathcal{W}}$ and $U_{\mathcal{W}}$ are bigram and unigram character-level language models built over all words in $\mathcal{W}$. We set $\lambda = 0.8$~\cite{IR-548} which was observed to be an empirically strong setting for GEN. We experimented with higher-order models in {\bf GEN}, but observed drops in evaluation measures leading to us sticking to the usage of the unigram and bi-gram models. The form of Eq.~\ref{eq:gen} is inspired by an assumption similar to that used in both~\cite{prasad2014technique} and~\cite{koo2015unsupervised} that loanwords are rare. Thus, we expect they would not be adequately supported by models that generalize over whole of $\mathcal{W}$; intuitively, $(1-w_n^{GEN})$ may be thought of as a score of outlierness, being an quantification of deviation from a model learnt over the corpus. We also compare against our diversity-based initialization score from Section~\ref{sec:divinit}, which we will call as {\bf INIT}. For ease of reference, we outline the {\bf INIT} scoring:

\begin{equation}
w_n^{INIT} = min\bigg\{ 0.99, \frac{|u3(w_{stem},\mathcal{W})|}{\tau}\bigg\}
\end{equation}

The comparison against {\bf INIT} enables us to isolate and study the value of the iterative update formulation vis-a-vis the initialization.

\subsection{Evaluation Outline}

We outlined in Section~\ref{sec:probdef}, we use {\it top-k}, {\it bottom-k} and {\it avg-k} precision (evaluated at varying values of $k$) as well as {\it clustering quality} in our evaluation. Of these, the clustering quality metric provides an overview of the performance of UNS vis-a-vis the baselines. The precisions at the ends of the ordering, allow for delving deeper into the orderings produced by the nativeness scores. Accordingly, we start by analyzing clustering quality of UNS across varying settings of $n$ (n-gram length), $\alpha$ (weighting between two terms in optimization) and $\rho$ (used in construction of highly diverse words set) against the baselines. This is followed by a similar analysis over the metrics of {\it top-k}, {\it bottom-k} and {\it avg-k} precisions against the baseline methods. We then perform a deeper analysis of UNS to understand its sensitivity over other parameters such as word stem length and $\tau$ (used in initialization), to conclude our empirical evaluation. Each of the above analyses are performed separately on the Malayalam and Telugu datasets described in Section~\ref{sec:datasets}. Unless mentioned otherwise, we set the word stem length to two, and the diversity threshold $\tau$ to $10$. UNS iterations were continued until there were fewer than 1\% of labels changing across successive iterations, or until $100$ iterations are reached, whichever is earlier. 

\subsection{Evaluation on Clustering Quality: UNS vs. Baselines}

Table~\ref{tab:dtim} and Table~\ref{tab:dtit} record the results of UNS against the baselines over the Malayalam and Telugu datasets respectively. As outlined in Section~\ref{sec:evaluationmeasures}, the tables list the clustering quality for the native and loanword clusters followed by the weighted average that provides a single evaluation measure over the entire dataset. The results over a wide variety of settings of $n$, $\alpha$ and $\rho$ suggest that UNS outperforms the baselines across a variety of parameter settings. Each of the UNS performance numbers on each of the measures can be seen to be better than the respective numbers for both the baselines. Of the varying parameter settings, $n=1$, $\alpha = 1.0$ and $\rho = 3$ consistently record the best numbers across both the Malayalam and Telugu datasets; the best numbers are boldfaced in the table. The technique peaking at the same parameter settings for both languages indicates that UNS modelling is able to exploit the commonalities in lexical structure between the two Dravidian languages. $n=1$ entails the usage of single character level probability distributions, whereas $\alpha = 1.0$ ensures an even weighting across the model conformance and inertia terms. $\rho$ marks the strength of the inertia in that smaller values of $\rho$ cause a larger set of words to be weighted into the inertia term; thus the higher performance of $\rho = 3$ as against $5$ further illustrates the importance of the inertia term in driving UNS towards desirable scoring. It is further notable that $\alpha = 0.0$, the setting that discards the inertia term, records a significant drop in performance as against settings where the inertia term is factored in. Overall, the clustering quality evaluation establishes the consistent performance of UNS and illustrates why UNS should be the preferred method for the task.

\begin{table*}
\centering
  \caption{UNS Clustering Quality for Malayalam vs. Baselines. (best numbers bold)}
  \label{tab:dtim}
\begin{tabular}{cccccc}
\hline
\hline
\multicolumn{3}{c}{Method} & Native & Loanword & Wt. Average \\
\hline
\multicolumn{3}{c}{INIT} & 0.79 & 0.38 & 0.69 \\
\multicolumn{3}{c}{GEN} & 0.73 & 0.17 & 0.59 \\
\hline
\hline
\multicolumn{6}{c}{UNS Results across Parameter Settings} \\
\hline
$n$ & $\alpha$ & $\rho$ & Native & Loanword & Wt. Average \\
\hline
1 & 0.0	& N/A & 0.838 &	0.512 &	0.756 \\
1 & 0.5 & 3	& 0.861 &	0.581 &	0.791 \\
1 & 0.5 & 5	& 0.860 &	0.578 &	0.789 \\
1 & 1.0 & 3	& {\bf 0.867} &	{\bf 0.601} &	{\bf 0.801} \\
1 & 1.0 & 5	& 0.858 &	0.574 &	0.787 \\

2 & 0.0	& N/A & 0.844 &	0.531 &	0.766 \\
2 & 0.5 & 3	& 0.861 &	0.581 &	0.791 \\
2 & 0.5 & 5	& 0.856 &	0.566 &	0.783 \\
2 & 1.0 & 3	& 0.863 &	0.589 &	0.795 \\
2 & 1.0 & 5	& 0.852 &	0.554 &	0.778 \\
\hline
\hline
\end{tabular}
\end{table*}

\begin{table*}
\centering
  \caption{UNS Clustering Quality for Telugu vs. Baselines}
  \label{tab:dtit}
\begin{tabular}{cccccc}
\hline
\hline
\multicolumn{3}{c}{Method} & Native & Loanword & Wt. Average \\
\hline
\multicolumn{3}{c}{INIT} & 0.80 & 0.40 & 0.70  \\
\multicolumn{3}{c}{GEN} & 0.72 & 0.17 & 0.59 \\
\hline
\hline
\multicolumn{6}{c}{UNS Results across Parameter Settings} \\
\hline
$n$ & $\alpha$ & $\rho$ & Native & Loanword & Wt. Average \\
\hline
1 & 0.0 & N/A & 0.889 &	0.665 &	0.834 \\
1 & 0.5 & 3 & 0.927 &	0.778 &	0.890 \\
1 & 0.5 & 5 & 0.918 &	0.751 &	0.876 \\
1 & 1.0 & 3 & {\bf 0.931} &	{\bf 0.790} &	{\bf 0.896} \\
1 & 1.0 & 5 & 0.918 &	0.751 &	0.876 \\
2 & 0.0 & N/A & 0.794 &	0.377 &	0.691 \\
2 & 0.5 & 3 & 0.806 &	0.412 &	0.708 \\
2 & 0.5 & 5 & 0.797 &	0.385 &	0.695 \\
2 & 1.0 & 3 & 0.810 &	0.424 &	0.714 \\
2 & 1.0 & 5 & 0.798 &	0.389 &	0.697 \\
\hline
\hline
\end{tabular}
\end{table*}

\begin{table*}
\centering
  \caption{Top-k, Bottom-k \& Average-k Results}
  \label{tab:endseval}
\begin{tabular}{l|ccc|ccc|ccc}
\hline
\multicolumn{10}{c}{{\bf Evaluation on Malayalam Dataset}} \\
\hline
& \multicolumn{3}{c|}{k = 50}  & \multicolumn{3}{c|}{k = 100}  & \multicolumn{3}{c}{k = 200} \\
\cline{2-10}
& Top-k & Bot-k & Avg-k & Top-k & Bot-k & Avg-k  & Top-k & Bot-k & Avg-k \\
\hline
INIT & {\bf 0.88} & 0.50 & 0.69 & {\bf 0.90} & 0.40 & 0.65 & {\bf 0.90} & 0.38 & 0.64 \\
GEN & 0.64 & 0.10 & 0.37 & 0.58 & 0.11 & 0.35 & 0.64 & 0.17 & 0.41 \\
\hline
UNS & 0.84 & {\bf 0.78} & {\bf 0.81} & 0.86 & {\bf 0.79} & {\bf 0.82} & 0.88 & {\bf 0.65} & {\bf 0.76} \\
\hline
\multicolumn{10}{c}{{\bf Evaluation on Telugu Dataset}} \\
\hline
& \multicolumn{3}{c|}{k = 50}  & \multicolumn{3}{c|}{k = 100}  & \multicolumn{3}{c}{k = 200} \\
\cline{2-10}
& Top-k & Bot-k & Avg-k & Top-k & Bot-k & Avg-k  & Top-k & Bot-k & Avg-k \\
\hline
INIT & {\bf 1.00} & 0.46 & 0.73 & {\bf 0.99} & 0.38 & 0.69 & {\bf 0.97} & 0.39 & 0.68 \\
GEN & 0.54 & 0.10 & 0.32 & 0.60 & 0.12 & 0.36 & 0.64 & 0.16 & 0.40 \\
\hline
UNS & 0.94 & {\bf 0.74} & {\bf 0.84} & 0.95 & {\bf 0.73} & {\bf 0.84} & 0.96 & {\bf 0.81} & {\bf 0.89} \\
\hline
\end{tabular}
\end{table*}

\subsection{Evaluation on End-Precisions: UNS vs. Baselines}

Table~\ref{tab:endseval} lists down the end-precision metrics, laid out in Section~\ref{sec:evaluationmeasures}, across varying values of $k$. While the top-k precision measures the fraction of {\it native} words at the native end of the ordering formed by the $w_n$ scores, bottom-k precision measures the fraction of {\it loanwords} at the other end. As may be obvious, what may be regarded as the key indicator is average-k precision, which forms the mean of the precision at either ends. It is however, to be noted that this evaluation only focuses on a subset of the dataset, and $|\mathcal{W}| - 2k$ data points are excluded from influencing the evaluation; thus, these evaluations only serve only the limited purpose of ensuring that the either ends of the ordering are {\it pure} in the expected sense. In many cases where automated scoring is applied to do a two-class classification, it may be desirable to subject the ambiguous band of objects in the middle to manual verification, whereas the labels in the end may be regarded as more trustworthy to bypass manual verification. Such manual verification may be inevitable in critical processes such as those in healthcare and security, making the end-precision measures being the more useful measure to analyze for such scenarios, as compared to clustering quality. Table~\ref{tab:endseval}, in the interest of brevity, lists the performance of UNS at the parameter setting $n=1$, $\alpha = 1.0$ and $\rho = 3$, the setting that was found to be most desirable for both langauges from the analysis in the previous section. It is easy to see from the results that UNS convincingly outperforms the baselines on the average-k measure across varying values of $k$ over both the languages; this confirms the observations from the previous section. It is interesting to note the trend of Top-k where INIT, the initiatialization used in UNS, scores better. Top-k measure the purity at the high-end of $w_n$ scores; this means that the initialization heuristic is very effective in putting the native words in the high $w_n$ range. However, it fares very badly in the low $w_n$ range, as indicated by the Bottom-k precision dropping to $<0.5$. This offers a perspective on UNS as well; starting from an ordering that is accurate only within the high $w_n$ territory, the UNS model is able to learn the probability distributions that are meaningful enough to spread the gains more evenly across the whole spectrum of $w_n$ scores over the course of the iterative refinements. 

\begin{table}
\centering
\begin{tabular}{c|ccccc}
		\hline
		\hline
		$\tau \rightarrow$ & 5 & 10 & 20 & 50 & 100 \\
		\hline
		\hline
		Malayalam & 0.89 & {\bf 0.90} & {\bf 0.90} & 0.58 & 0.55 \\
		\hline
		Telugu & 0.78 & {\bf 0.80} & {\bf 0.80} & 0.77 & 0.75 \\
		\hline
		\end{tabular}
\caption{UNS Clustering Quality with Varying $\tau$}
\label{tab:tau}
\end{table}

\begin{table}
\centering
\begin{tabular}{c|cccc}
		\hline
		\hline
		$Stem\ Length \rightarrow$ & 1 & 2 & 3 & 4 \\
		\hline
		\hline
		Malayalam & 0.63 & 0.80 & {\bf 0.82} & 0.59 \\
		\hline
		Telugu & 0.67 & {\bf 0.90} & 0.78 & 0.52 \\
		\hline
		\end{tabular}
\caption{UNS Clustering Quality with Varying Stem Length}
\label{tab:stem}
\end{table}

\subsection{UNS Evaluation: Parameter Sensitivity and Objective Function Trends}

\noindent{\bf $\tau$ and Word Stem Length:} Up until above, we retained the word stem length to be consistently two, and the diversity threshold parameter $\tau$ was set to $10$ across all analyses. In this section, we study the sensitivity of UNS to these two parameters. All other parameters are set to values as earlier. The results across varying values of $\tau$ and stem length are shown in Tables~\ref{tab:tau} and~\ref{tab:stem} respectively. The table suggests that UNS is extremely robust to variations in diversity threshold, despite a slight preference towards values around $10$ and $20$. This suggests that a system designer looking to use UNS need not carefully tune this parameter due to the inherent robustness. Given the nature of Malayalam and Telugu where the variations in word lengths are not as high as in English, it seemed very natural to use a word stem length of $2$. Moreover, very large words are uncommon in Malayalam and Telugu. In our corpus, $50\%$ of words were found to contain five characters or less. Our analysis of UNS across variations in word-stem length, illustrated in Table~\ref{tab:stem} strongly supports this intuition with clustering quality peaking at stem-length close to $2$ (for Malayalam, the actual peak is at $3$, but the clustering quality improvement for $3$ over $2$ is not much). It is notable, however, that UNS degrades gracefully beyond that range. Trends across different settings of word-stem length are interesting since they may provide clues about applicability for other languages with varying character granularities (e.g., each Chinese character corresponds to multiple characters in Latin-script).

\begin{figure}[tb]
\centering
\includegraphics[width=0.8\columnwidth]{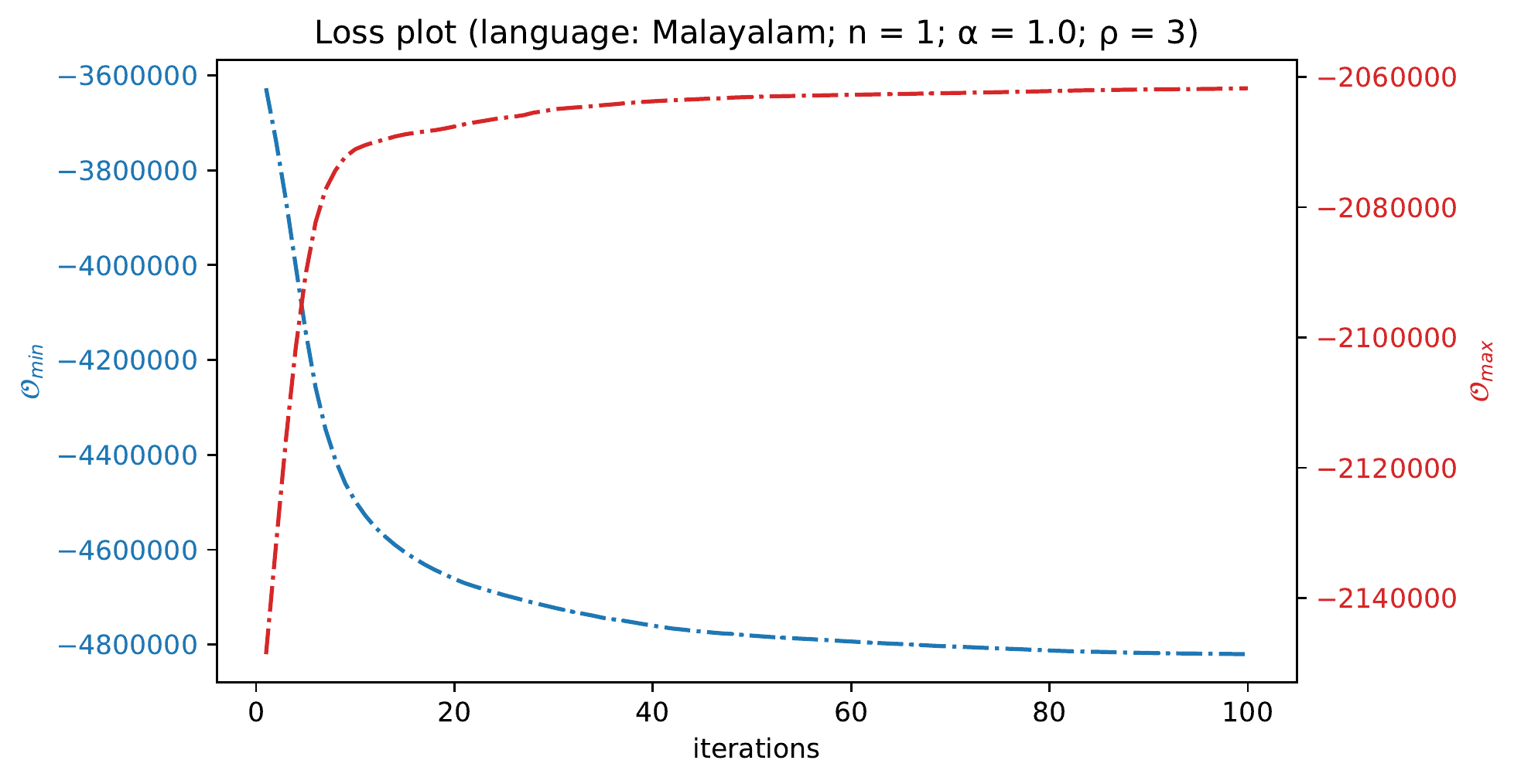}
\caption{Minimizing and Maximizing Objective Function Values (Left and Right Y-axes respectively) vs. Iterations (X-Axis) for Malayalam}
\label{fig:objectivesml}
\end{figure}

\begin{figure}[tb]
\centering
\includegraphics[width=0.8\columnwidth]{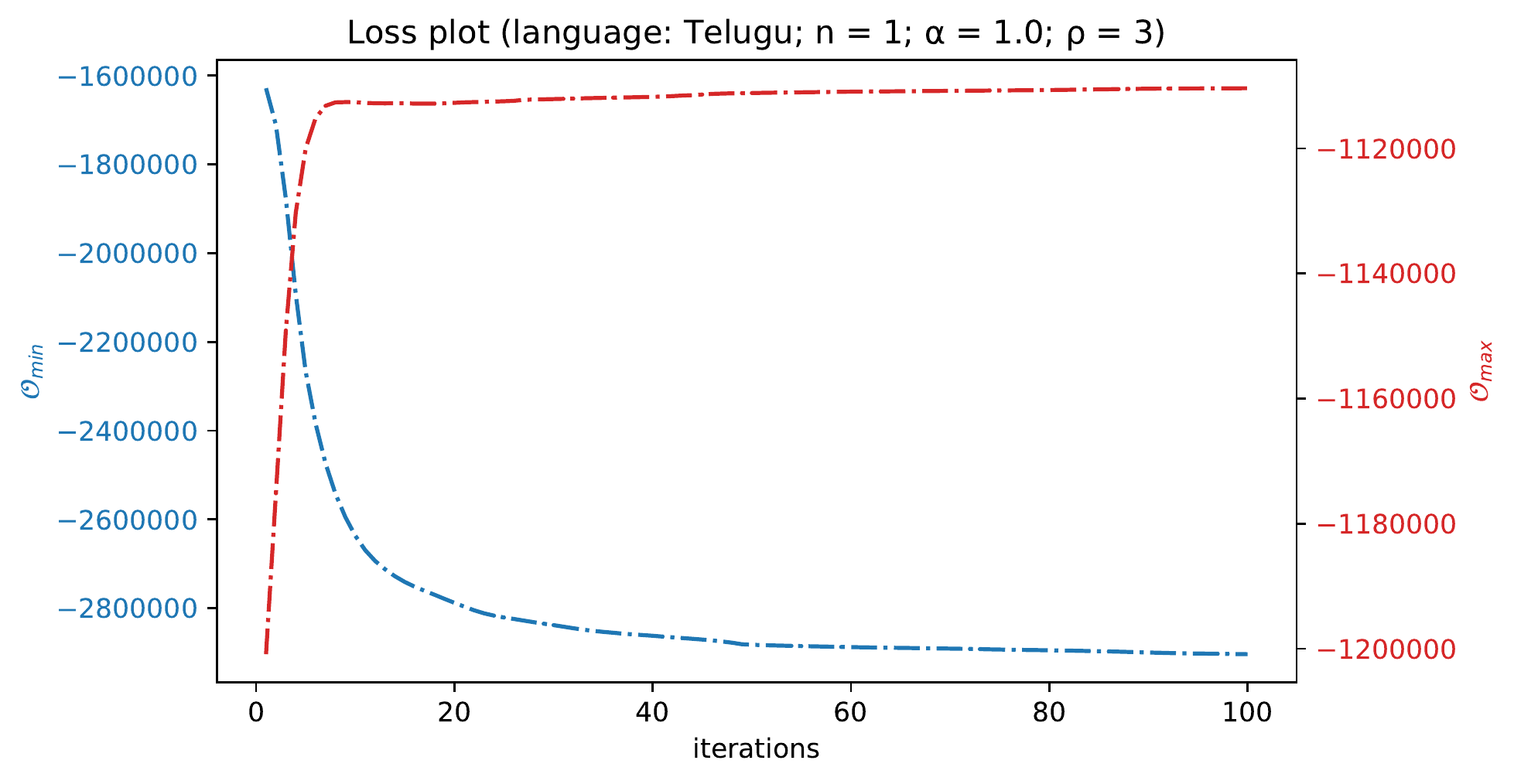}
\caption{Minimizing and Maximizing Objective Function Values (Left and Right Y-axes respectively) vs. Iterations (X-Axis) for Telugu}
\label{fig:objectivestl}
\end{figure}

\noindent{\bf Objective Functions Across Iterations:} Across UNS iterations, we expect the maximizing and minimizing objectives to be progressively refined in appropriate directions. Figure~\ref{fig:objectivesml} plots the objective function trends across $100$ iterations for the Malayalam dataset when UNS is run with $n=1, \alpha = 1.0, \rho = 3$. The trends, as expected, show rapid objective function changes (max objective function increasing and min objective function decreasing, as expected) in the initial iterations, with the values stabilizing beyond $20$ to $30$ iterations. Similar trends were observed for the Telugu dataset as well as for varying settings of hyperparameters; the corresponding chart appears in Figure~\ref{fig:objectivestl}.. That the objective functions show a steady movement towards convergence as iterations progress, we believe, is indicative of the effectiveness of the UNS formulation. 

\begin{figure}[tb]
\centering
\includegraphics[width=0.8\columnwidth]{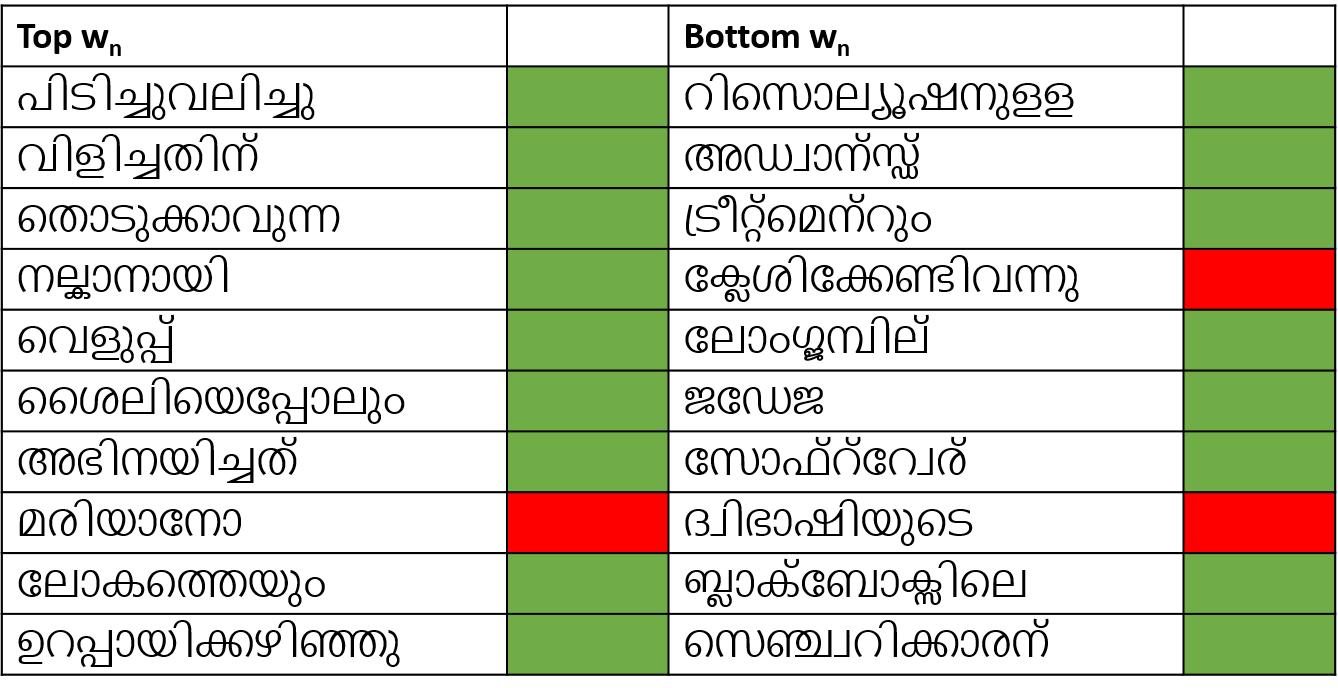}
\caption{Top-10 {\it Malayalam} words from either ends of the $w_n$ spectrum. The words at the top $w_n$ ($1.0$ end) are shown along with their labels as either native (green) or non-native (red). For the bottom $w_n$ end ($0.0$ end), the label colorings are flipped. }
\label{fig:maltopk}
\end{figure}

\begin{figure}[tb]
\centering
\includegraphics[width=0.8\columnwidth]{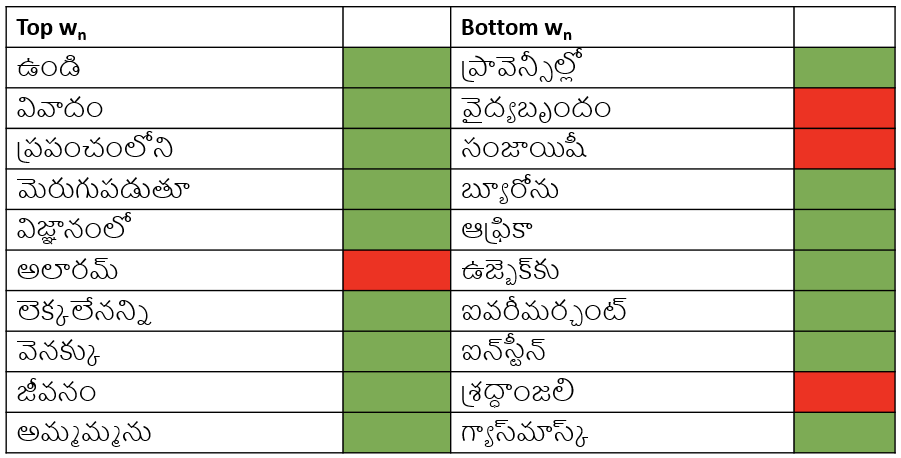}
\caption{Top-10 {\it Telugu} words from either ends of the $w_n$ spectrum. The words at the top $w_n$ ($1.0$ end) are shown along with their labels as either native (green) or non-native (red). For the bottom $w_n$ end ($0.0$ end), the label colorings are flipped. }
\label{fig:teltopk}
\end{figure}

\subsection{UNS Qualitative Analysis}

Towards analyzing the results qualitatively, we now present the top-10 words on either ends of the $w_n$ spectrum for Malayalam and Telugu in Figures~\ref{fig:maltopk} and~\ref{fig:teltopk} respectively. The labelling is driven by the motivation to illustrate {\it correctness} which depends on the choice of ends; this is detailed in the caption of the respective figures. Based on an analysis, we found that the highest values of $w_n$ are generally achieved for words that are commonly used, with the native words that appear at the low $w_n$ end being those that are quite rarely used.

\section{Discussion}

\subsection{Applicability to Other Languages} 

In contrast to earlier work focused on specific languages (e.g.,~\cite{koo2015unsupervised}) that use heuristics that are very specific to the language (such as expected patterns of consonants), the UNS framework is general-purpose in design. The main heuristic setting that is likely to require some tuning for applicability in other languages, such as other Indic languages, is the word-stem length. We expect the approach would generalize well to other Sanskrit-influenced Dravidian languages such as Kannada, Tulu and Kodava, but may require some adaptation for others such as Tamil due to a lack of diversity in the alphabet. Unfortunately, we did not have any Kannada/Tulu/Kodava knowledge (Dravidian languages have largely disjoint speaker-populations) in our team, or access to labelled datasets in those languages (they are low-resource languages too); testing this on Kannada/Tulu/Tamil would be interesting future work. 

\subsection{The Nuanced Nature of Word Nativity}

As an empirically oriented work, we have considered native and non-native as two distinct and different concepts. This is reflected in our formulation of $w_n$ as a nativeness score and $1-w_n$ as a loanwordness score. This is also reflected in our evaluation dataset that makes use of binary native/non-native labels. However, as may be well-understood, {\it nativeness} is a much more nuanced concept. A loanword that has been in usage for a long time in a language may be regarded as native for every practical purpose, making the mutual exclusivity embedded in the $w_n/1-w_n$ construction obsolete. For example, $/ka/se/ra$, a widely used malayalam-language word to denote {\it chair}, has its origins in the portugese word {\it cadeira}; with the embedding of the word $/ka/se/ra$ within Malayalam being so pervasive to the extent that most native speakers are unaware of the portugese connection, it may be argued to have both high nativeness and high loanwordness. Additionally, langauges such as Sanskrit that have influenced some dravidian languages for many centuries, have contributed words that take part in productive and complex morphological processes within the latter. For this and other reasons, it may be meaningful to consider a more structured scoring and labelling process for words in extending UNS to scenarios that would need to be sensitive to such distinctions. 

\subsection{UNS in an Application Context} 

Within any target application context, and especially so in domains such as healthcare and security, machine-labelled non-native words (and their automatically generated transliterations) may need to manual screening for accountability reasons. The high accuracy at either ends of the ordering lends itself to be exploited in the following fashion; in lieu of employing experts to verify all labellings/transliterations, low-expertise volunteers (e.g., students/Mechanical Turkers) can be called in to verify labellings at the ends (top/bottom) of the lists with experts focusing on the middle (more ambiguous) part of the list; this frees up experts' time as against a cross-spectrum expert-verification process, leading to direct cost savings. 


\section{Conclusions and Future Work}\label{sec:conclusions}

We considered the problem of unsupervised separation of loanwords and native words in Malayalam and Telugu; this is a critical task in easing automated processing of Malayalam/Telugu text in the company of other language text. We outlined a key observation on the differential diversity beyond word stems, and formulated an initialization heuristic that would coarsely separate native and loanwords. We proposed the usage of probability distributions over character n-grams as a way of separately modelling native and loanwords. We then formulated an iterative optimization method that alternatively refines the nativeness scorings and probability distributions. Our technique for the problem, UNS, that encompasses the initialization and iterative refinement, was seen to significantly outperform other unsupervised baseline methods in our empirical study. This establishes UNS as the preferred method for the task. We have also released our datasets and labeled subset to help aid future research on this and related tasks. 


\subsection{Future Work}

The applicability of UNS to other Indic languages is interesting to study. Due to our lack of familiarity with any other language in the family, we look forward to contacting other groups to further the generalizability study. While nativeness scoring improvements directly translate to reduction of effort for manual downstream processing, quantifying gains these bring about in translation and retrieval is interesting future work. Exploring the relationship/synergy of this task and Sandhi splitting~\cite{DBLP:conf/ijcnlp/NatarajanC11} would form another interesting direction for future work. 

Loanwords are often within Malayalam used to refer to very topical content, for which suitable words are harder to find. Thus, loanwords could be preferentially treated towards building rules in interpretable clustering~\cite{balachandran2012interpretable} and for modelling context in regex-oriented rule-based information extraction~\cite{murthy2012improving}. Loanwords might also hold cues for detecting segment boundaries in conversational transcripts~\cite{kummamuru2009unsupervised,padmanabhan2007mining}. 

\section{Acknowledgements}
The authors would like to thank Giridhara Gurram for annotating the Telugu dataset.




\bibliography{ms_bib}
\end{document}